\documentclass{article}

\usepackage[nonatbib, preprint]{neurips_2024}
\PassOptionsToPackage{numbers}{natbib}

\usepackage[utf8]{inputenc} % allow utf-8 input
\usepackage[T1]{fontenc}    % use 8-bit T1 fonts
\usepackage{url}            % simple URL typesetting
\usepackage{booktabs}       % professional-quality tables
\usepackage{amsfonts}       % blackboard math symbols
\usepackage{nicefrac}       % compact symbols for 1/2, etc.
\usepackage{microtype}      % microtypography
\usepackage{xcolor}         % colors
\usepackage{graphicx}
\usepackage{hyperref}       % hyperlinks
\usepackage{algorithm}
\usepackage{algpseudocode}

\hypersetup{colorlinks=true, breaklinks=true, %pagebackref=true,
  urlcolor=blue, linkcolor=blue,anchorcolor=blue,citecolor=blue,
  hypertexnames=true, final=true, 
  pdfpagemode = UseNone, %FullScreen, %UseThumbs, %UseOutlines,
  pdfauthor = {},
  pdftitle = {},   
  pdfsubject = {},
  pdfkeywords = {}
}

\graphicspath{{../figures/} {figures/} {./}}
\DeclareGraphicsExtensions{.pdf,.png,.jpg,.mps}

\title{Neuromorphic dreaming: A pathway to efficient learning in artificial agents}

\author{%
  Ingo Blakowski$^{1,2}$, Dmitrii Zendrikov$^1$, Cristiano Capone$^{3,*}$, Giacomo Indiveri$^{1,*}$\\
  $^1$Institute of Neuroinformatics, University of Zurich and ETH Zurich\\
  $^2$Technical University of Munich\\
  $^3$Natl. Center for Radiation Protection and Computational Physics,\\ Istituto Superiore di Sanità, 00161 Rome, Italy\\
  $^*$shared last author\\
  \texttt{\{ingoblakowski,cristiano0capone\}@gmail.com}\\ 
  \texttt{\{dmitrii,giacomo\}@ini.uzh.ch} \\
}

\begin{document}

\maketitle

\begin{abstract}
Achieving energy efficiency in learning is a key challenge for artificial intelligence (AI) computing platforms.
Biological systems demonstrate remarkable abilities to learn complex skills quickly and efficiently.
Inspired by this, we present a hardware implementation of model-based reinforcement learning (MBRL) using spiking neural networks (SNNs) on mixed-signal analog/digital neuromorphic hardware.
This approach leverages the energy efficiency of mixed-signal neuromorphic chips while achieving high sample efficiency through an alternation of online learning, referred to as the "awake" phase, and offline learning, known as the "dreaming" phase.
The model proposed includes two symbiotic networks: an agent network that learns by combining real and simulated experiences, and a learned world model network that generates the simulated experiences.
We validate the model by training the hardware implementation to play the Atari game Pong. 
We start from a baseline consisting of an agent network learning without a world model and dreaming, which successfully learns to play the game.
By incorporating dreaming, the number of required real game experiences are reduced significantly compared to the baseline.
The networks are implemented using a mixed-signal neuromorphic processor, with the readout layers trained using a computer in-the-loop, while the other layers remain fixed.
These results pave the way toward energy-efficient neuromorphic learning systems capable of rapid learning in real world applications and use-cases.

\end{abstract}

\section{Introduction}
The field of artificial intelligence (AI) has been relentlessly developing more advanced and powerful models.
Recently, the introduction of transformer architectures and state-space models, combined with the strategy of scaling them to unprecedented sizes have led to groundbreaking achievements.
These advancements have been primarily driven by the steady progress in digital chip technology, facilitating the parallel computation of immense artificial neural networks on large-scale clusters.

However, midst the enthusiasm surrounding the ever-growing neural networks implemented on digital chips, the critical aspect of energy efficiency is often overlooked.
In contrast, biological systems demonstrate the remarkable ability to learn complex skills quickly and efficiently, often with limited experience and data.

\paragraph{Reinforcement learning with spiking neural networks.}
In an attempt to build more biologically plausible models, deep reinforcement learning (DRL) algorithms, such as Deep Q-Network (DQN) and Twin-Delayed Deep Deterministic Policy Gradient (TD3), have been adapted for spiking networks, which are used in both discrete and continuous action space environments~\cite{patel2019improved,tang2021deep,akl2023toward}.
These adaptations demonstrate the potential of spiking networks to handle complex control problems using advanced DRL techniques, enhancing their suitability for energy-efficient execution on neuromorphic processors.
Spike-driven processing, weight updates, and communication are particularly beneficial for reducing energy consumption on neuromorphic hardware~\cite{zenke2021remarkable}.

Nonetheless, most training algorithms used depend on non-local learning rules, e.g.\@ combining backpropagation with surrogate gradients, which are computationally intensive and biologically implausible~\cite{zenke2021remarkable}.
Designing powerful and efficient learning methods, with local rules that are also biologically plausible, and therefore better suited to being mapped onto energy efficient neuromorphic hardware platforms, remains an open challenge~\cite{khacef2023local}.
There are methods that focus on biologically plausible learning rules and architectures, such as recurrent spiking networks~\cite{florian2007reinforcement,fremaux2013reinforcement,bellec2020}.
These methods include reward-based local plasticity rules, which work well for simple tasks but face limitations in complex control scenarios~\cite{fremaux2013reinforcement}.
The e-prop method, a biologically plausible form of actor-critic and backpropagation through time, represents a state-of-the-art approach in spiking network RL, achieving comparable performances to non-spiking systems in benchmarks like Atari games~\cite{bellec2020,stockl2021optimized}.
Despite these advancements, efficiently utilizing limited online data remains a significant challenge.

To address this challenge, we use a recently proposed model-based reinforcement learning (MBRL) approach~\cite{capone2022towards} that uses spiking neural networks (SNNs) and is compatible with neuromorphic hardware implementations. This method is demonstrated to be more sample-efficient than state of the art model-free reinforcement learning (RL) approaches for spiking networks~\cite{bellec2020solution}.
The compatibility of this MBRL approach with neuromorphic hardware offers several advantages, which will be discussed further in the following section.

\paragraph{Neuromorphic hardware.}
Neuromorphic computing systems aim to emulate the computational principles of biological neural networks using specialized hardware substrates.
These systems often employ analog or mixed-signal circuits to efficiently implement neuron and synapse dynamics, enabling low-power and scalable implementations of spiking neural networks~\cite{mead1990neuromorphic, indiveri2011neuromorphic}.

In this work we are using the general-purpose neuromorphic processor architecture DYNAP-SE~\cite{moradi2017scalable}, which implements essential neuronal dynamics with the exponential leaky integrate-and-fire (ExLIF) model.
The neuron and synapse circuits exhibit biologically realistic temporal dynamics with time constants spanning from microseconds to hundreds of milliseconds~\cite{chicca2014neuromorphic}.
To minimize dynamic power consumption and achieve biophysically plausible behaviors, the analog neuron and synapse circuits in the DYNAP-SE operate in the sub-threshold domain.
However, this design choice introduces sensitivity to device mismatch effects, resulting in heterogeneity in the individual neuron and synapse characteristics~\cite{pavasovic1994characterization}.
The chip area optimizations of the DYNAP-SE architecture also impose certain constraints, such as a limited number of neurons per core and chip, a single globally shared set of parameters (synaptic weight, time constants, etc.) per core, a maximum of 64 pre-synaptic connections per neuron, and sensitivity to temperature variations and other environmental factors.

Techniques to mitigate the impact of device mismatch and environmental variations, such as population coding, on-chip learning and calibration mechanisms, could enhance the robustness and reliability of neuromorphic systems~\cite{zendrikov2023brain}.
Building computational systems that robustly operate within these constraints benefits the field of in-memory computing as a whole, as non-negligible levels of variability are characteristic of many other classes of energy-efficient devices, such as memristors~\cite{querlioz2013immunity, indiveri2013integration, payvand2019neuromorphic}.

By combining the aforementioned sample efficiency of the MBRL approach and the compatibility with energy-efficient neuromorphic hardware implementations, we developed a hardware MBRL system that can run on-line in real-time, using an energy-efficient device with sub-milliWatt power consumption figures~\cite{moradi2017scalable}.
Due to its mixed-signal event-based nature the spiking neurons implemented in this device only consume power when they are active (i.e. when they receive input spikes).

The key contributions of this work are twofold:
\begin{enumerate}
    \item We present a real-time implementation of the MBRL spiking neural network on neuromorphic hardware.
    \item We validate our approach by demonstrating state-of-the-art performance on the Atari Pong benchmark, achieving sample-efficient learning with limited interactions with the environment.
\end{enumerate} 
The remainder of this paper is structured as follows: 
Section \ref{sec:methodology} presents our neuromorphic model-based RL approach, detailing the spiking network architecture, learning rules, and mapping to neuromorphic hardware.
Section \ref{sec:experiments} describes our experimental setup and presents empirical results on the Atari game Pong.
Sections \ref{sec:discussion} and \ref{sec:conclusion} conclude our research and discuss the implications of our findings, the limitations of our approach, and suggests directions for future research.

\section{Methodology}\label{sec:methodology}
We first explore how reinforcement learning can be implemented using spiking neural networks.
We use a model-based approach similar to \cite{capone2022towards} with two networks: an agent network for decision making, that produces action probabilities to select actions based on the current state of the environment, and a model network for simulating the environment dynamics, which means predicting the change of the environment state and the next reward.
In the following sections, we present the network architectures and their corresponding plasticity rules, along with a detailed explanation of how each network generates its output. We also describe the "awake-dreaming" approach used in our model-based reinforcement learning system. Additionally, we explain the techniques employed for encoding input states into spiking neural network (SNN) population codes and interpreting the output spikes.

% \paragraph{Reinforcement learning with spiking neural networks:}
Several approaches have been proposed for implementing reinforcement learning in spiking networks.
One key idea is to modulate spike-timing-dependent plasticity based on a reward signal~\cite{bi1998synaptic,florian2007reinforcement}.
Building upon these ideas, we use a biologically plausible approach~\cite{capone2022towards} for reinforcement learning in spiking networks that draws inspiration from the reward-based e-prop method~\cite{bellec2020solution}.
In contrast to modulated STDP rules, our approach does not modify input and recurrent connections.
Instead, we focus on training only the readout weights that connect the spiking neurons to the output layer, which is implemented on a computer interacting with the neuromorphic chip~\cite{moradi2017scalable}.
This choice is motivated by the constraints of the neuromorphic hardware, which uses the same input synapse weight for all neurons and allows a maximum number of incoming connections per neuron~\cite{moradi2017scalable}.
By restricting plasticity to the readout connections, we can work within these limitations while still enabling the network to learn from rewards.

\subsection{Agent network}

The agent's policy $\pi^t_k$ is defined as the probability of selecting action $k$ at time $t$.
We represent this probability using a softmax function applied to the output activations $y^t_k$ of the readout layer, similar to what is done in \cite{capone2022towards}:
\begin{equation}
\pi^t_k = \frac{exp(y^t_k)}{\sum_i exp(y^t_k)}
\end{equation}
where $y^t_k$ is computed as a weighted sum of the filtered spiking activity $\bar{s}^t_{\alpha,i}$ from the hidden layer neurons:
\begin{equation}
y^t_k = \sum_i R^\pi_{ki} \bar{s}^t_{\alpha,i}
\end{equation}

\paragraph{Learning rule:}
The objective is to maximize the cumulative reward over time.
We formalize this using a loss function $E^A$, similar as in \cite{capone2022towards,bellec2020solution}:
\begin{equation}
E^A = -\sum_t R^t log(\pi^t_k)
\end{equation}
where $R^t = \sum_{t' \geq t} \gamma^{t'-t} r^{t'}$ is the discounted return, $\gamma$ is the discount factor, and $r^t$ is the reward received at time $t$.
To optimize the policy, we update the readout weights $R^\pi_{ik}$ using a learning rule inspired by the policy gradient component of reward-based e-prop~\cite{capone2022towards,bellec2020solution}:
\begin{equation}
\Delta R_{ik}^\pi = -\eta_\pi \sum_t r^t \sum_{t' \leq t} \gamma^{t-t'} (\pi_{k}^{t'} - 1_{a^{t'}=k}) \bar{s}_{A,i}^{t'}
\end{equation}
Here, $\eta_\pi$ is the learning rate, and $1_{a^{t'}=k}$ is an indicator function that equals 1 when the chosen action $a^{t'}$ is $k$ and 0 otherwise.
This learning rule is driven by the difference between the actual action and the probability assigned by the policy, weighted by discounted rewards.
The filtered spiking activity $\bar{s}^{t'}_{\alpha,i}$ serves as an eligibility trace, indicating the contribution of each hidden neuron to the selected action.
By updating the readout weights proportionally to this reward-weighted difference, the agent's policy is gradually improved to maximize cumulative reward.
While our approach does not modify input and recurrent weights as in the full reward-based e-prop algorithm, it still enables effective learning of state-to-action mappings through plasticity of the readout connections, while respecting hardware constraints~\cite{moradi2017scalable}.

\paragraph{Network architecture:}
Our reinforcement learning model employs a feed-forward spiking network with a hidden layer.
The input layer encodes state variables using spike trains generated by on-chip spike generators.
We use $m_{agent}=40$ spike generators, with $10$ dedicated to each state variable.
These feed into a hidden layer of $n_{agent}=510$ leaky integrate-and-fire (LIF) silicon neurons implemented on the chip~\cite{moradi2017scalable}.
The input-to-hidden connections are randomly initialized and remain fixed during learning.
This allows us to work within the constraints of the hardware, which has limited flexibility in modifying these connections.
The hidden neurons integrate incoming spikes and generate output spike patterns based on their membrane potential dynamics.
Their spiking activity is forwarded to an artificial neural network readout layer implemented on the computer.
The hidden-to-readout connections are learned using the reward-based policy gradient rule, which adjusts synaptic weights to maximize expected cumulative reward.
The readout layer consists of $k_{agent}$ neurons, where $k_{agent}$ is the number of possible actions.
For Atari Pong, we have $k_{agent}=3$ readout neurons corresponding to the actions "Up", "Down", and "Stay".
The readout layer applies a softmax function to the incoming activations, producing a probability distribution over available actions.
The action with the highest probability is then selected for execution.

\subsection{World model network}
We propose a biologically plausible learning scheme for the world model using spiking networks on the neuromorphic chip, similar to the agent network.
We use e-prop~\cite{bellec2020solution} but in a supervised fashion.
Training focuses only on the readout connections, while leaving input-to-hidden weights fixed.

The objective is to predict the next state $\xi_k^{t+1}$ and reward $r^{t+1}$ given the current state $\xi_k^{t}$ and action $a^t$ selected by the agent.
The state and reward predictions are defined as linear readouts of the spiking activity in the model network~\cite{capone2022towards,bellec2020solution}: 
\begin{equation}
    \xi_k^{t+1} = \sum_i R_{ki}^\xi \bar{s}^t_{M,i}
\end{equation}
\begin{equation}
    r^{t+1} = \sum_i R^r_i \bar{s}^t_{M,i}
\end{equation}
Here, $R_{ki}^\xi$ and $R^r_i$ are the readout weights for the state and reward predictions, and $\bar{s}_{M,i}^t$ represents a low-pass filtered version of the spike rates from the model network hidden neurons.
\paragraph{Learning rules:}
To train the world model, we minimize the loss function~\cite{capone2022towards,bellec2020solution}:
\begin{equation}
E^M = c_\xi \sum_{t,k} (\xi^{\star t+1}_k - \xi^{t+1}_k)^2 + c_r \sum_t (r^{\star t+1} - r^{t+1})^2
\end{equation}
where $\xi^{\star t+1}_k$ and $r^{\star t+1}$ are the target next state and reward, and $c_\xi$ and $c_r$ are coefficients balancing the state and reward prediction errors.
Applying e-prop yields the following update rules for the readout weights~\cite{capone2022towards,bellec2020solution}:
\begin{equation}
\Delta R^\xi_{ik} = -\eta_\xi \sum_t (\xi^{\star t+1}_k - \xi^{t+1}_k) \bar{s}^t_{M,k}
\end{equation}
\begin{equation}
   \Delta R^r_k = -\eta_r \sum_t (r^{\star t+1} - r^{t+1}) \bar{s}^t_{M,k}
\end{equation}
These local learning rules enable online training of the world model readout weights~\cite{capone2022towards}.
\paragraph{Network architecture:}
The world model employs a feedforward spiking network consisting of an input layer, a hidden layer of silicon neurons, and a readout layer.
The input layer encodes the current state variables and selected action using spike trains generated by on-chip spike generators.
We dedicate $m_{model}=40$ spike generators to the state input, with $10$ allocated to each state variable.
The action input is encoded using $k_{model}=3$ spike generators, one for each possible action in Pong.
The state and action input spike trains are transmitted to a hidden layer of $n_{model}=510$ silicon LIF neurons on the chip.
The input-to-hidden synaptic connections are randomly initialized and remain fixed during learning, allowing the hidden layer to serve as a reservoir of temporal features.
The spiking activity of the hidden neurons is projected to the readout layer, which resides on the computer, via learnable synaptic weights.
The readout layer consists of $s_{model}=4$ artificial neurons for predicting the next state and one neuron for estimating the expected reward.
The state and reward readout weights are updated using the supervised learning rules derived from e-prop, enabling the world model to capture environment dynamics.

\subsection{Awake-dreaming learning}
\begin{figure}[t!]
  \centering
  \label{fig:dreaming}
  \includegraphics[width=.95\linewidth]{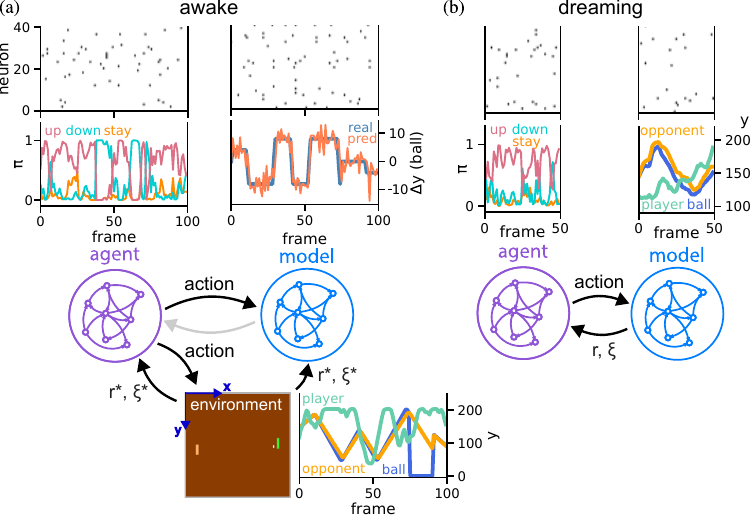} 
  % \fbox{\rule[-.5cm]{0cm}{4cm} \rule[-.5cm]{4cm}{0cm}}
  \caption{Training with dreaming alternates between two phases.
\textbf{(a)} In the awake phase, consisting of 100 frames, the agent network (policy) and model network interact with the real environment.
The agent takes the current real state and predicts an action, which is performed in the game and fed into the model network.
The real reward is used to compute the policy gradient, while the model network predicts the change in state variables and reward, which are compared to the actual state and reward to update the model frame by frame.
\textbf{(b)} In the dreaming phase, lasting 50 frames, the agent network is updated by interacting solely with the model network, detached from the real environment.
The agent takes the imaginary current state from the world model and predicts an action, which is fed into the world model to predict the next state and reward.
The imaginary reward is used to compute the policy gradient for updating the agent network.}
\end{figure}

Inspired by the role of dreaming in memory consolidation and reinforcement learning in biological brains~\cite{stickgold2001sleep, walker2004sleep}, we adopt an "awake-dreaming" approach~\cite{capone2022towards}. 
The schematic of the approach is shown in Figure \ref{fig:dreaming}.
During the awake phase, the agent interacts with the environment, updating both its policy and the world model based on the observed transitions and rewards.
In the dreaming phase, the agent is disconnected from the environment and instead uses the learned world model to generate simulated experiences, which are then used to further refine the policy.
This alternation between real and imagined experiences allows the agent to learn more efficiently by leveraging the sample-efficiency of model-based methods while still benefiting from the generalization capabilities of model-free techniques.
The dreaming phase also enables the agent to continue learning even when real-world interactions are not possible or too costly.

\subsection{Input Encoding}
To interface the spiking neural network with the Atari Pong environment, we need to transform the game state variables and the chosen action into spike trains that can be processed by the networks.
We employ a population coding technique \cite{Deneve_etal01}, where the index of the most active population encodes the value of the encoded variable.
\paragraph{Environment State Encoding:}
The Atari Pong game state consists of four variables: the positions of the two paddles, the ball's x-coordinate, and the ball's y-coordinate.
For each variable, we define a population of input spike generators that respond differently to specific values or positions (Figure \ref{fig:input_encoding}).
As the value of a state variable changes, the Gaussian activity profile moves across the population of input spike generators associated with that variable.
This encoding scheme allows the network to effectively represent and process the game's continuous state space.
\paragraph{Action Encoding:}
In addition to the game state variables, the chosen action must also be encoded and provided as input to the world model network.
As there are three possible actions in Atari Pong (move paddle up, down, or stay), we employ a separate set of input spike generators to represent the selected action.

To enable the world model network to effectively process the action information, the action input spike generators are connected to the entire population of hidden layer neurons using random weights.
These connections are quantized, meaning the weights can assume a discrete set of values.
In our implementation, we use different numbers of parallel connections between the action spike generators and the hidden neurons to represent these quantized weights.
For instance, a stronger weight would be represented by a higher number of parallel connections, while a weaker weight would have fewer connections.

\section{Experiments and results}\label{sec:experiments}

\begin{figure}[!htbp]
  \centering
  \label{fig:setup_and_timing}
  \includegraphics[width=1.0\linewidth]{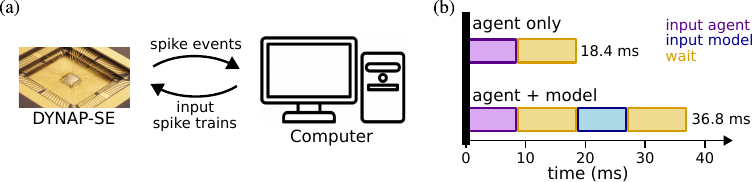}
  \caption{Setup and timing diagram.
\textbf{(a)} The setup employs a computer-in-the-loop system, where the DYNAP-SE neuromorphic chip efficiently simulates the neural dynamics of the input and hidden layers, while the computer handles chip-environment synchronization and manages the learning protocol.
The readout layers are implemented on the computer, and learning focuses on updating the output weights stored on the computer, while the input-to-hidden connections on the chip remain fixed due to hardware constraints.
The computer generates input spike trains based on the game state and action, which are loaded onto the chip.
After a short waiting period for the chip to process the new input, the computer reads out the spike events, which are then used as input (in the form of the number of spikes per neuron) to the readout layers to predict the respective outputs.
\textbf{(b)} The timing diagram illustrates the duration of one step with and without dreaming.
The majority of the time is consumed by updating the input to the agent or model and the subsequent waiting period, while other processing times are negligible.
Updating the input takes 8.4 ms, and the waiting period is 10 ms.
Consequently, a step without the model takes 18.4 ms, while a step with the model takes 36.8 ms.}
\end{figure}

% \paragraph{Experimental setup:}
To evaluate our neuromorphic implementation, we conducted a series experiments on the Atari Pong video game environment from the OpenAI Gym toolkit~\cite{brockman2016openai}.
Our implementation builds upon the code provided by \cite{capone2023biodreamingcode}, which we adapted to incorporate our neuromorphic techniques to interact with the chip in real-time using the Python API of the Samna software \cite{Samna}.

Our primary aim was to assess the sample efficiency of the approach, measured by the average return achieved per game as a function of the number of frames of experience.
We also tracked the entropy of the agent's policy over the course of training to visualize how the action selection confidence evolves.

Each training run consists of 2000 games, with each game lasting 100 frames.
This setup resulted in a total of 200,000 frames of experience per training run, providing a comprehensive assessment of the agent's learning progress.
To account for potential variability in performance, we conducted 10 independent runs for each experimental configuration and reported the averaged results.
This approach ensures the robustness and reliability of our findings, enabling us to draw meaningful conclusions.

\subsection{Setting neural parameters}
Before starting our training runs, we needed to set the neural parameters of the DYNAP-SE neuromorphic chip.
Finding the right setting includes a combination of adjusting the synaptic weight, neuron/synapse time constants, and impulse response strength.
Often, it is sufficient to only adjust the synaptic weight such that all the neurons in the network have a certain integration factor, which is defined as:
\begin{equation}
\textnormal{integration factor} = \frac{\textnormal{\#output events}}{\textnormal{\#incoming events}}
\end{equation}
We can compute this integration factor by counting the number of output events for one game and dividing it by the number of incoming events during that game.
In our experiments, an integration factor between $0.45$ and $0.58$ has proven to work well.

\subsection{Baseline agent without dreaming}
We first established a baseline using an SNN-based agent without any dreaming capabilities.
The baseline agent architecture consists of an input layer using population coding to represent the game state, a hidden layer of $510$ leaky integrate-and-fire neurons implemented on the DYNAP-SE chip, and a 3-unit readout layer corresponding to the 3 actions in Pong.
Only the readout weights are updated during training, using the reward-based policy gradient rule.
Figure \ref{fig:setup_and_timing} \textbf{(a)} depicts the overall setup, with the neuromorphic chip handling the neural dynamics and the computer managing the input/readout interfaces and learning.
Figure \ref{fig:setup_and_timing} \textbf{(b)} (top) illustrates the timing breakdown, showing a total of 18.4 ms per step to process a single frame and update the agent.
To optimize the policy readout we used the Adam optimizer \cite{kingma2014adam} together with a learning rate of $0.004$.

\subsection{Agent with dreaming}
We next experimented with augmenting the agent to incorporate dreaming, using a separate model network which learns the environment dynamics.
The model network has a similar architecture to the agent, but has 3 additional action inputs and $4 + 1$ readout units to predict the next state and reward.
Training alternates between "awake" phases, where both the agent and model learn from $100$ frames of real experience, and "dreaming" phases, where the agent learns from $50$ steps on imagined trajectories sampled from the model.
The timing breakdown in Figure \ref{fig:setup_and_timing} \textbf{(b)} (bottom) shows that dreaming increases the per-frame training time to $36.8$ ms.
We used learning rates of $0.002$ for the policy and state readouts and $0.0004$ for the reward readout. The policy readout weights are initialized by sampling from a normal distribution $\mathcal{N}(0,0.1)$, while the state and reward readout weights are initialized with $0$.
 
\subsection{Timing considerations}
Optimizing system performance and training time required careful consideration of the waiting period between updating the input and reading out spikes from the hidden neurons.
We investigated waiting times of 10 ms, 20 ms, and 50 ms, observing no significant changes in performance.
Consequently, we selected a 10 ms waiting time to minimize training time while maintaining performance.
Figure \ref{fig:setup_and_timing} \textbf{(b)} illustrates the timing breakdown of agent steps in two training scenarios.
\paragraph{No dreaming:}
The agent-only scenario using only awake learning takes 18.4 ms per step (8.4 ms for updating input, 10 ms for waiting), resulting in 1.84 s for one training game (100 frames).
Training the agent network for 2000 games without a model network takes approximately 1.02 hours.
\paragraph{Dreaming:}
The scenario incorporating both awake and dreaming phases takes 36.8 ms for a combined agent and model step (Two times 8.4 ms for updating input, two times 10 ms for waiting).
This results in a training game duration of 5.52 s (100 real frames + 50 dreaming frames).
Training the agent and model networks for 2000 games with both phases takes around 3.07 hours.
The majority of step time is dedicated to updating spike generators and the waiting period.
To balance performance and training time, we opted for fewer spike generators and a shorter waiting time, allowing for satisfactory performance while keeping training times reasonable.

\subsection{Results}
\begin{figure}[!htbp]
  \centering
  \label{fig:results}
  \includegraphics[width=1.0\linewidth]{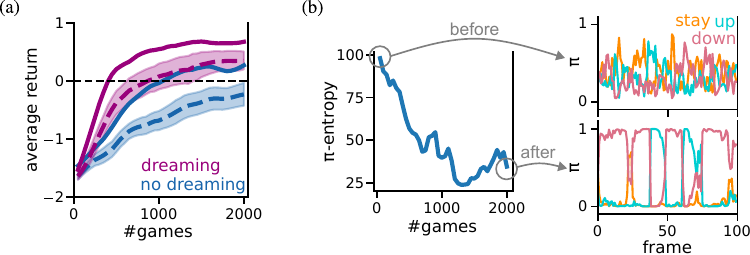} 
  \caption{\textbf{(a)} Average return per game over the last 50 games as a function of the number of games played, for an agent that uses dreaming (purple) and a baseline that does not (blue).
The dashed line represents the mean over 10 independent training realizations, the solid line represents the 80th percentile, and the shaded area represents the standard deviation.
With dreaming, the average return increases significantly faster in terms of interactions with the real environment.
\textbf{(b)} Evolution of the policy entropy during training, quantifying the uncertainty in action selection at each game.
Before training (top right), the policy shows high uncertainty, while after training (bottom right), the policy adapts quickly and assigns high probabilities to single actions, indicating growing confidence as learning progresses.}
\end{figure}

The core results are presented in Figure \ref{fig:results}.
Panel \textbf{(a)} compares the average return achieved by the agent with and without dreaming over the course of training.
We observe that the incorporation of dreaming leads to a significant increase in sample efficiency on Pong, allowing the agent to reach higher scores with fewer than half as many real environment frames. 
In addition, with dreaming, the agent seems to be more robust against getting stuck at "low-return" policies. This can be better seen from the separately plotted training curves in the supplementary material, where the agent escapes low returns quite fast and reliably throughout independent training runs, compared to the training runs without dreaming.
Panel \textbf{(b)} visualizes the evolution of the policy entropy for a representative run with dreaming.
The agent gradually becomes more confident which is shown by the decreasing policy entropy over training time.
Useful behaviors are quickly identified and reinforced.

\section{Discussion}
\label{sec:discussion}
\subsection{Contributions and implications:}
To our knowledge, this is the first time the DYNAP-SE neuromorphic processor has been used with an approach that updates the input spike trains in a millisecond time frame, enabling real-time interaction with the environment.
This represents a significant achievement in utilizing neuromorphic chips for real-world applications.

These findings have implications for the development of sample-efficient and energy-efficient learning systems.
The biologically-inspired approach we use, together with the computational advantages of our neuromorphic implementation, offers a promising direction for creating intelligent agents that can learn and adapt in real-world settings with limited data and power consumption.

\subsection{Limitations}
Despite the promising results, our work has several limitations.
First, due to constraints of training time and network size, we only trained the agent for a short time horizon, decreasing the complexity of the task.
Second, because of the limitation of having only one synaptic weight for all connections on a core and the maximum of 64 incoming connections per neuron, we were unable to train the networks directly on the chip.
Instead, learning was restricted to the readout connections.
Another challenge we encountered was the variability between different DYNAP-SE chips.
The neuron and synapse parameters, in combination with the connection configuration and the learned readout weight matrix, that worked well on one chip, usually do not yield the same performance on another chip, likely due to device mismatch and temperature changes~\cite{zendrikov2023brain}. %Additionally, changes in the timing of spike readouts led to differences in spike behavior and a temporary drop in performance during training.
Moreover, as the complexity of the tasks and environments increase, training the world model may become more challenging.
The model network might require more diverse and variable information about the environment to accurately capture its dynamics.

\subsection{Future directions}
There are several promising directions for future research based on our findings.
One direction is to explore the transfer of the readout layers to the neuromorphic chip by quantizing the weights and using parallel connections or by leveraging next-generation chips, which offer more programmable features and synaptic weights.
These advancements could enable the solution of more complex tasks using neuromorphic hardware.

Another potential option for future work is the utilization of Poisson spike generators for input encoding.
With further engineering optimizations on the DYNAP-SE chip, the update time for Poisson spike generators could be significantly reduced.
This would allow for a more biologically plausible input representation while maintaining the system's real-time interaction capabilities.
Investigating the performance of our approach with optimized Poisson spike generators could yield valuable insights into the role of neural coding.

Another important step is to further test our approach on a wider range of tasks, including more complex games and real-world applications.
This will help assess the generalizability and scalability of our approach running on neuromorphic hardware.
To address the challenge of training the world model for more complex tasks and environments, it could be beneficial to employ multiple agents during training.
By using different agents to gather more variable information about the environment, the model network can be exposed to a wider range of experiences, potentially improving its ability to capture the environment's dynamics.

\section{Conclusion}\label{sec:conclusion}
In this work, we have presented the implementation of a model-based reinforcement learning approach using spiking neural networks, on neuromorphic hardware.
We have verified the suitability of the DYNAP-SE, a low-power optimized neuromorphic chip, for our approach.
Despite the constraints and challenges associated with analog neuromorphic hardware, we successfully trained an agent to play the game Pong, achieving good performance with reduced environmental interactions by leveraging a "dreaming" phase.
In conclusion, our work demonstrates the potential of model-based reinforcement learning with spiking networks on neuromorphic hardware for creating sample-efficient and energy-efficient learning systems.
By drawing inspiration from biological neural networks and leveraging the computational advantages of neuromorphic chips, we can develop intelligent agents that learn and adapt in real-world environments with limited data and power consumption.
With the introduction of next-generation neuromorphic chips, we anticipate that more complex tasks can be tackled using this approach.
Future research in this direction can lead to significant advances in neuromorphic computing and its applications in robotics, autonomous systems, and beyond.

{\small \bibliography{main_paper}}

%%%%%%%%%%%%%%%%%%%%%%%%%%%%%%%%%%%%%%%%%%%%%%%%%%%%%%%%%%%%
\clearpage
\appendix

\section{Appendix}
% Optionally include supplemental material (complete proofs, additional experiments and plots) in appendix.
% All such materials \textbf{SHOULD be included in the main submission.}

\subsection{Algorithm}
\begin{algorithm}
\caption{Pseudocode for the neuromorphic dreaming algorithm.}\label{ALG:dreaming}
\begin{algorithmic}[!htbp]
\While{$iteration < N_{iter}$}\\
\State \textbf{Awake phase}
\For{$T_{awake} = 100$ iterations}
\Comment{play one \textbf{real} game}
\State{perform one action in the real env.}
\State update readout parameters of the model-network $\{R_{ki}^{\xi},R_{i}^{r}\}$
\Comment{supervised}
\EndFor
\State update readout parameters of the agent-network $\{R_{ki}^A\}$
\Comment{policy gradient}\\
\State \textbf{Dreaming phase}
\For{$T_{dream} = 50$ iterations}
\Comment{play one \textbf{simulated} game}
\State{perform one action in the simulated env.}
\EndFor
\State update readout parameters of the agent-network $\{R_{ki}^A\}$
\Comment{policy gradient}\\
\EndWhile
\end{algorithmic}
\end{algorithm}
The agent alternates between an awake phase, where it interacts with the real environment and updates the readout parameters of the world model network using supervised learning, and a dreaming phase, where it "dreams" by playing out simulated experiences using the learned world model and further optimizes its policy using the policy gradient method in this simulated environment. The hyperparameters $T_{awake}$ and $T_{dream}$ control the number of iterations spent in each phase.

\subsection{Network architectures}
\label{supp:network_architecture}
\begin{figure}[!htbp]
  \centering
  \label{fig:networks}
  \includegraphics[width=1.0\linewidth]{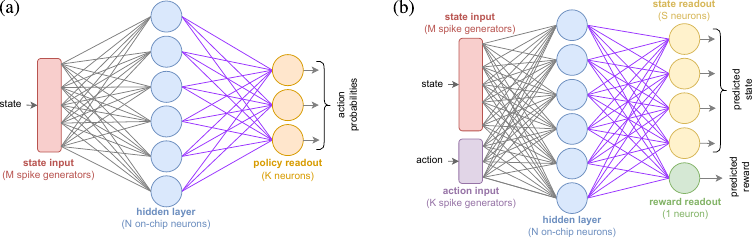} 
  \caption{Network architectures of the agent and model networks.
  (a) The agent network predicts action probabilities based on the state input encoded by population-coded spike generators. The input-to-hidden connections (gray) are fixed and randomly initialized such that each hidden neuron receives exactly 8 incoming connections. These connections are assigned one of four quantized synaptic weights, chosen randomly, which are implemented by 1 to 4 parallel connections between an input neuron and a hidden neuron. The learnable hidden-to-output connections (purple) are trained using a policy gradient method on the computer in the loop with the neuromorphic chip.
  (b) The model network predicts the next state and reward based on the current state and action inputs. The fixed input-to-hidden connections (gray) follow a similar random connectivity pattern as the agent network. The hidden-to-output connections (purple) are learned in a supervised manner to minimize state and reward prediction errors.}
\end{figure}

\subsection{Input encoding}
Some examples of how the state values are encoded and transformed into spike trains are shown in Figure 5.
\begin{figure}[!htbp]
  \centering
  \label{fig:input_encoding}
  \includegraphics[width=1.0\linewidth]{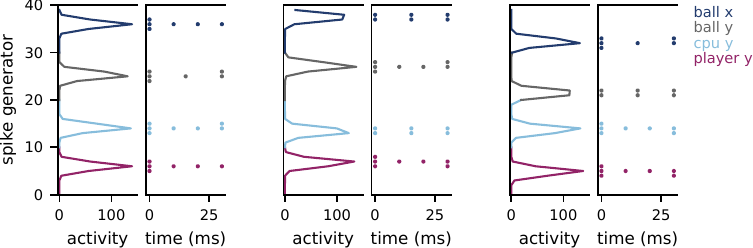} 
  \caption{Population coding for the four state variables in Atari Pong. The activity of the input spike generators follows a Gaussian distribution, with the mean shifting according to the corresponding state variable's value between the minimum and maximum spike generator index of the corresponding population.}
\end{figure}

\section{Additional details on experimental settings}
\subsection{Hardware}
All experiments have been conducted on a single server that is connected to several DYNAP-SE1 boards. The relevant specifications of the server and the DYNAP-SE1 boards are listed below:

Server:
\begin{itemize}
    \item CPU: Intel(R) Core(TM) i7-6700K CPU @ 4.00GHz
    \item RAM: 50 GB 
\end{itemize}
DYNAP-SE1 board:
\begin{itemize}
    \item 4 x DYNAP-SE chips (interconnectable, each has 4 cores with 256 neurons)
    \item One single shared set of parameters for all neurons/synapses on one core
    \item 4 types of synapses (AMPA, NMDA, GABA A, GABA B)
\end{itemize}

\subsection{Hyperparameters}
More details on the hyperparameters are given in Table \ref{tab:hyperparams}.
\begin{table}[!htbp]
\caption{Hyperparameters for training.}
\label{tab:hyperparams}
\centering
\begin{tabular}{lll}
\toprule
\textbf{Hyperparameter}     & \textbf{Value} \\
\midrule
\multicolumn{2}{c}{\textit{General}} \\
\midrule
Number of games per training run & 2000 \\
Number of independent training runs & 10 \\
Frames in awake phase ($T_{awake}$) & 100 \\
Frames in dreaming phase ($T_{dream}$) & 50 \\
Discount factor ($\gamma$) & 0.998 \\
\midrule
\multicolumn{2}{c}{\textit{Agent only}} \\
\midrule
Policy learning rate ($\eta_\pi$)       & $4$ x $10^{-3}$     \\
Number of hidden neurons ($n_{agent}$)     & $510$      \\
Number of input spike generators ($m_{agent}$)     & $40$      \\
\midrule
\multicolumn{2}{c}{\textit{Agent + model}} \\
\midrule
Policy learning rate ($\eta_\pi$)       & $2$ x $10^{-3}$     \\
State prediction learning rate ($\eta_\xi$)       & $2$ x $10^{-3}$     \\
Reward prediction learning rate ($\eta_r$)       & $4$ x $10^{-4}$     \\
Number of hidden neurons in agent ($n_{agent}$)     & $510$      \\
Number of hidden neurons in model ($n_{model}$)     & $510$      \\
Number of input spike generators for agent ($m_{agent}$)     & $40$      \\
Number of input spike generators for model ($m_{model}$)     & $40 + 3$      \\
\bottomrule
\end{tabular}
\end{table}

\subsection{DYNAP-SE parameters}
Upon request, the authors can provide parameters for the DYNAP-SE. These parameters may serve as a starting point and offer guidance on the appropriate range for the parameters.

\section{Additional training curves}
This section presents individual training curves for three distinct experiments. Figure 6 illustrates an experiment conducted without dreaming and utilizing a lower learning rate. Figure 7 demonstrates the outcomes of another non-dreaming experiment but employs a higher learning rate. Lastly, Figure 8 depicts a training run that incorporates dreaming while maintaining the same lower learning rate as in Figure 6.
\begin{figure}[!htbp]
  \centering
  \label{fig:train1}
  \includegraphics[width=1.0\linewidth]{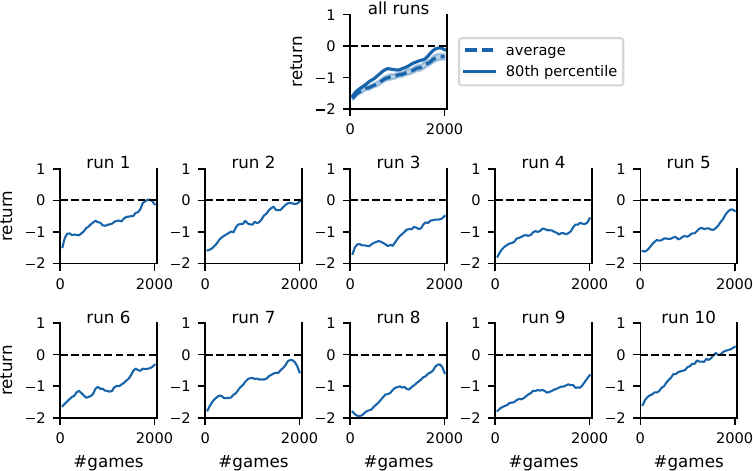} 
  \caption{Training curves of our baseline with \textbf{no dreaming} for 10 independent training runs with the following learning rate: $\eta_\pi$ = $2$ x $10^{-3}$.}
\end{figure}
\begin{figure}[!htbp]
  \centering
  \label{fig:train2}
  \includegraphics[width=1.0\linewidth]{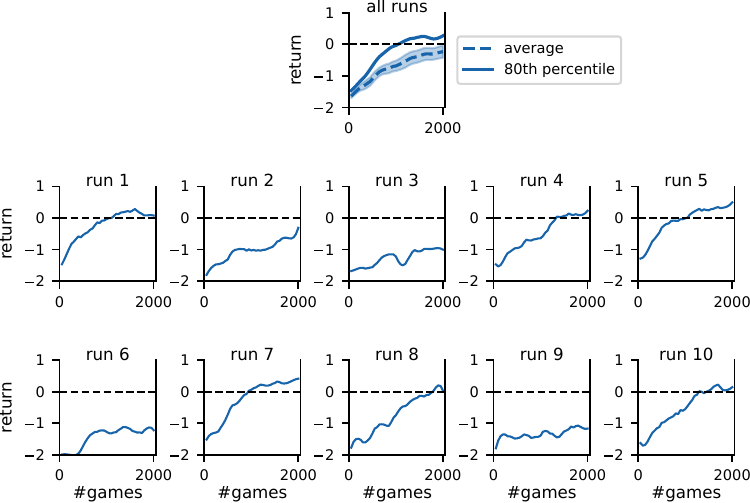} 
  \caption{Training curves of our baseline with \textbf{no dreaming} for 10 independent training runs with the following learning rate: $\eta_\pi$ = $4$ x $10^{-3}$.}
\end{figure}
\begin{figure}[!htbp]
  \centering
  \includegraphics[width=1.0\linewidth]{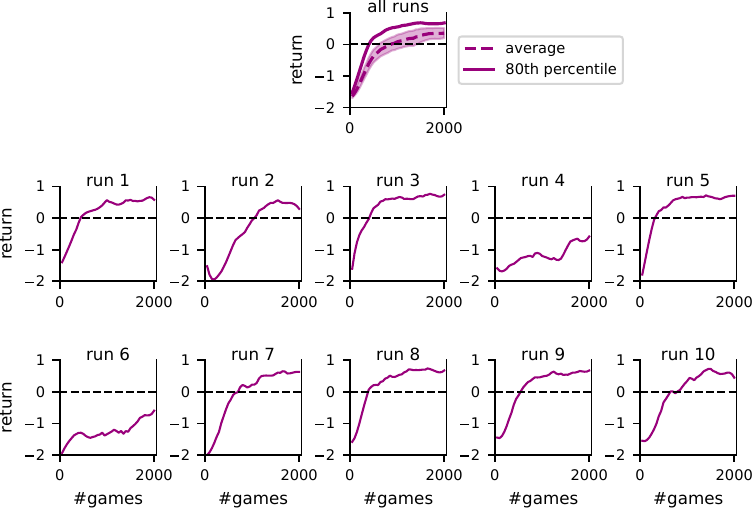} 
  \caption{Training curves of our baseline with \textbf{dreaming} for 10 independent training runs with the following learning rates: $\eta_\pi$ = $2$ x $10^{-3}$, $\eta_\xi$ = $2$ x $10^{-3}$, $\eta_r$ = $4$ x $10^{-4}$.}
\end{figure}

\section{Video}
Upon request, the authors can provide a video demonstrating our neuromorphic agent playing Pong. The video shows the agent's performance at the beginning and at the end of training, visualizing the network's spiking activity, the learned policy, and the received rewards.

\section{Source code}
The source code is available under \url{https://github.com/blakeyy/neuromorphic_dreaming}.

%%%%%%%%%%%%%%%%%%%%%%%%%%%%%%%%%%%%%%%%%%%%%%%%%%%%%%%%%%%%
% Checklist
% \include{Styles/neurips_checklist}

\end{document}